# LANE DETECTION FOR PROTOTYPE AUTONOMOUS VEHICLE

Sertap Kamçı, Dogukan Aksu*, Muhammed Ali Aydin

Computer Engineering Department, Istanbul University-Cerrahpasa, Istanbul, Turkey
1306150012@ogr.iu.edu.tr,
d.aksu@istanbul.edu.tr *,
aydinali@istanbul.edu.tr

## ABSTRACT

*Unmanned vehicle technologies are an area of great interest in theory and practice today. These technologies have advanced considerably after the first applications have been implemented and cause a rapid change in human life. Autonomous vehicles are also a big part of these technologies. The most important action of a driver has to do is to follow the lanes on the way to the destination. By using image processing and artificial intelligence techniques, an autonomous vehicle can move successfully without a driver help. They can go from the initial point to the specified target by applying pre-defined rules. There are also rules for proper tracking of the lanes. Many accidents are caused due to insufficient follow-up of the lanes and non-compliance with these rules. The majority of these accidents also result in injury and death.*

*In this paper, we present an autonomous vehicle prototype that follows lanes via image processing techniques, which are a major part of autonomous vehicle technology. Autonomous movement capability is provided by using some image processing algorithms such as canny edge detection, Sobel filter, etc. We implemented and tested these algorithms on the vehicle. The vehicle detected and followed the determined lanes. By that way, it went to the destination successfully.*

## KEYWORDS

*Autonomous Vehicle, Lane Detection, Image Processing, HSV Color, RGB Color, Canny Edge Detection, DC Motor, Region of Interest (ROI), Vanishing Point, Sobel Filter*

## 1. INTRODUCTION

Lane tracking is one of the most important tasks for autonomous vehicles. There are one or more lanes on each road. In a vehicle without lane tracking, there is no steering and this causes to accidents.

Prior to the development of autonomous vehicles, the vehicles had Active Lane Tracking Assistance. This system is used by many famous vehicle brands due to the lack of autonomous vehicles. The application of this system includes different types. These are just sound and vibration warning, steering interventions instead of the driver etc. This system helps keep the vehicle in the lane. Therefore, the importance of lane tracking system is understood.

A driving experience without lane tracking is not considered. In order to move with the correct timing, it must follow the path in real motion while in motion. This monitoring is usually performed in accordance with the timing of the images taken from the camera.

The algorithm of the vehicle allows to protect the lane. It adjusts the vehicle speed according to the lane. In addition, with the designed algorithm, the autonomous vehicle gives the vehicle an idea of what kind of moves it should follow if it wishes to cross a different lane in the future. First, the lane where the autonomous vehicle is located has to be detected and the snapshots are captured from the camera. These snapshots are called frames. There are many methods of image processing and are used to detect the lane. The lane lines are with the help of the camera which can see both sides of the vehicle. First, the quality of the picture should be improved and the lines should be determined. After that, the vanishing point and the Region of Interest (ROI) are determined. With the determined ROI, the vehicle understands how the lane continues in the direction. With the determination of the lane, the vehicle is given the instructions to continue straight to the vehicle according to the angles determined by the lane and to turn right or left.

This paper is organized as follows: Section 2 contains relevant studies and provides a comprehensive overview of the different detection methods that are used to the project. The functions and models used in Chapter 3 are presented. The experience gained in Chapter 4 is explained and the project output is given. In Chapter 5, the operating logic of the car following the lane is explained with a flow diagram. The results obtained are given in Chapter 6.

## 2. RELATED WORKS

The interest in autonomous vehicles has been increasing rapidly in recent years. In this paper, image processing algorithms for lane detection and tracking are mentioned.

There are several techniques for performing lane detection. First of all, the image processing methods used in the project should be examined to find the lines. Thereafter, lines should be identified from image frames where image processing techniques are applied. After this step, the ROI between the lines should be found. In addition, a virtual line is drawn in the middle of these lines to make the path look like a curve. With this curve, the bend of the way is determined by finding the angle. With this angle, the vehicle has an angle of movement and the vehicle moves in a straight, right-handed, left-handed direction according to the specified angle.

In this paper, we examine the related studies in three stages. These steps are: image processing techniques, detecting lane lines and finding ROI.

### 2.1. Image Processing Techniques

In the design of the algorithm, firstly the mutation from RGB color space to HSV color space is made. According to some studies, it is more convenient to use HSV color space when we want to differentiate an object of a certain color in any computer vision/image processing application [1]. In addition, the HSV color space provides clarity of the colors in the image. HUE is also used to distinguish colors more clearly.

Color images contain more information than gray level images. For this reason, in some related studies, the edge detection process for color images has been examined. Grayscale method is one of these methods. The cost of detecting ROI in the image was reduced by the Grayscale image conversation method and the process was accelerated [2,3].

### 2.2. Determination of Lane Lines

Canny edge detection, Hough transformation and Sobel filter methods are the methods used to find the lane lines.

Canny edge detection is a method of edge detection. The lanes are lines and have edges. Hough transformation method is a method that finds and shows shapes. Since the lane lines have a shape, lane lines can be found by the Hough transform method [4].

Sobel filter method is a separate method used to find the edge.

It is seen in all the studies obtained the use of the edge detection algorithm in images with a grayscale colour space is closer to the edge information in the actual image.

### 2.3. Finding Region of Interest (ROI)

The area of the lane is called the ROI. While some algorithms have an ROI up to the point where the lines can be detected by combining the lane lines in the image, some algorithms have methods for finding the ROI based on vanishing point detection techniques [5].

There are two common methods for the detection of ROI. Firstly, left and right lane lines are found. In the first method, the lane lines are stretched and intersect at one point. A triangular region is formed. The area within the triangular region up to the region where the lane are detected forms a rectangular region. This field returns the ROI. The deviation difference of the angles and the obtained region is calculated and the direction of the path is calculated. [1]

Another method is to determine the ROI by determining the skyline. Horizon line is the line where the earth globe and sky intersect when viewed in nature. With the determination of the horizon line, the lane lines are also combined and the intersecting area shows the ROI.

## 3. DESIGN OF THE LANE FOLLOWING AUTONOMOUS CAR

The design part of the car is divided into two parts. These are the movement of the software and the hardware of the vehicle. The software field is algorithm design and coding.

### 3.1. Software

The designed code has a layered architecture. There is a python file with a main structure named Main. Methods in other Python files are imported and run by calling the methods respectively. The lane lines must be determined in the algorithm of the car following the lane. For this, lane detection algorithm is used. Image processing techniques are used in the lane detection algorithm. These image processing techniques are described in the below.

By applying image processing techniques, lane lines are determined from the obtained image. These lane lines are calculated and centered in the algorithm and center lane is formed. The algorithm written for ROI detection is called here and the calculation of the ROI is done in this algorithm. The ROI-finding algorithm maintains the intersecting and polygon definition region on the image. This is the area between the lane and the horizon. In order to detect the lanes, image processing methods are applied to the given picture frames of the frame name taken from the image in the received camera. These methods; Conversion from RGB to HSV color space, Gaussian Blur, Gray Scale method, Canny Edge Detection, Hough Transform methods were used.

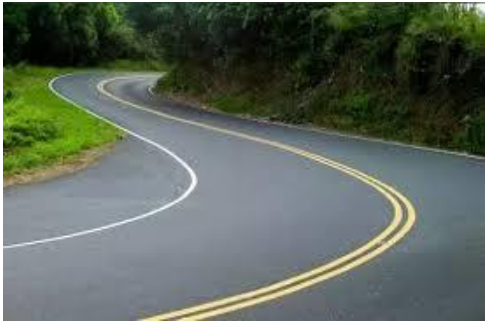

(a) Original frame

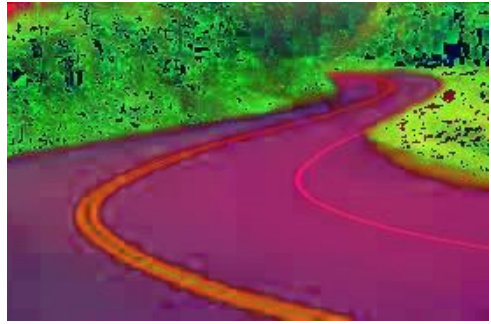

(b) HSV frame

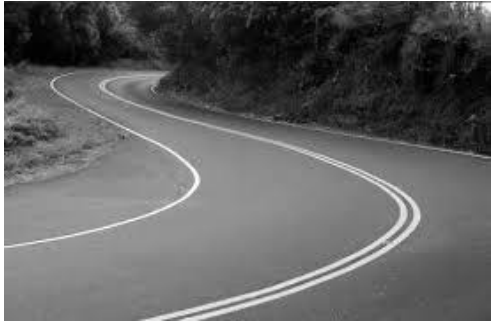

(c) Grayscale frame

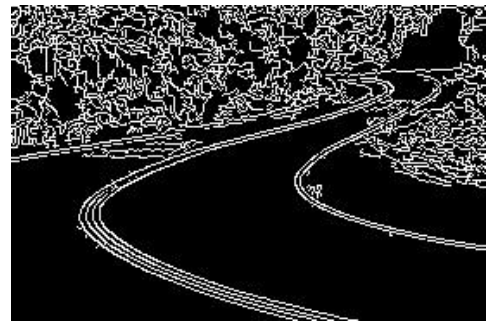

(d) Canny Edge Frame

Figure 1.  Steps of image processing methods on a path

In Figure 1 (a), a path with RGB color space was converted to HSV color space as in Figure 1 (b). Color saturation increases with HSV color space. Then, the image was processed as shown in Figure 1 (c) using Gaussian Blur and Gray Scale methods. Figure 1 (d) is used to understand the line of the lanes on the road by using Canny edge detection and Hough Transform methods. As a result of the calculations, the strip lines are perceived as shown. In Figure 2-a, a re-converted image frame is given to the RGB image. In Figure 2-b, an image frame that is not converted to RGB color space is given.

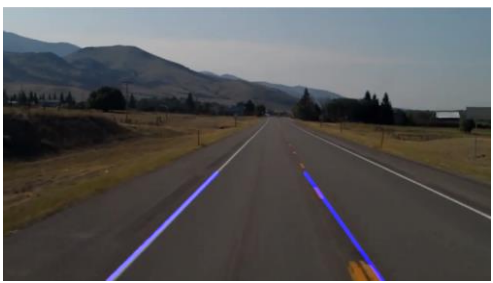

(a) Original path image

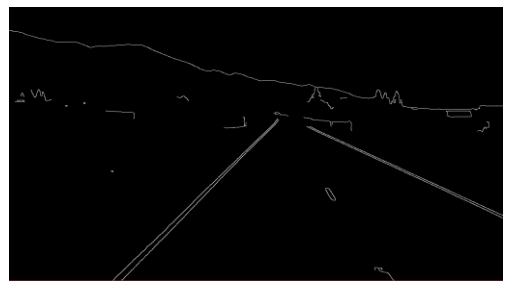

(b) Detected Lane from image

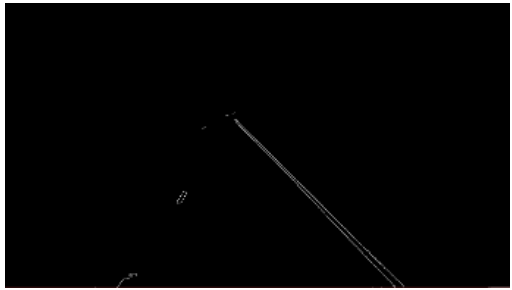
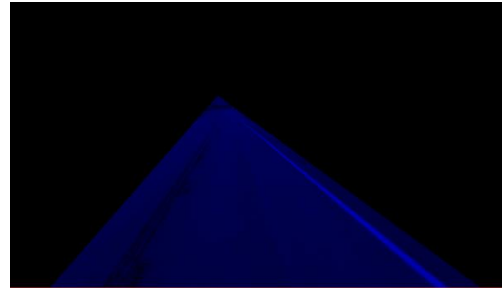

        (c) Detected interested lane        (d) ROI of the path image

Figure 2. Detection of lanes in a sample path

In Figure 2 (a), the algorithm is written to detect the road where the vehicle is located and this lane is shown based on the nearest lane. In Figure 2 (b), the ROI of the lane was found.

In Figure 2-c, the algorithm is written to detect the path where the vehicle is located and this lane is shown based on the nearest lane. Figure 2-d shows the ROI of the lane.

### 3.2. The Movement of the Vehicle

We want the vehicle to center the strips before start moving. In the code, the function that generates the virtual line is written to center. Then you need to follow the lane where it is located. For tracking of the lane, the ROI has the angle of rotation of the lane. When driving straight ahead, the vehicle is commanded to continue straight ahead. We used DC motors for the movement of the vehicle.
GPIO.output(Motor1A,GPIO.LOW)
Above left is the left motor stop command. In the same way, the right engine must be stopped for a right turn. This stop command may be slightly different, if the vehicle is not going to turn completely and rotating at a different degree than the angle of 180, the motor in the direction of rotation is decelerated and the motor in the other direction is accelerated. This is the logic of the vehicle's progress and rotation relative to the road.

### 3.3. Hardware

Raspberry Pi and Raspberry Pi camera are used as hardware. Robot car kit, breadboard, cables, six battery bed and power bank were used for the operation and integration of the vehicle. In order to keep the power of the vehicle uninterrupted, the six-bed battery was connected to the motors and the power bank was started to power the raspberry pi. The final version of the vehicle is shown in Figure 3.

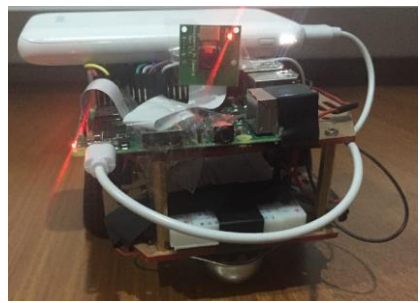

Figure 3. Image of autonomous vehicle prototype

## 4. EXPERIMENTS

The most important part was the progress of the coding step by step. During the last construction phase of the vehicle, the movement of the vehicle in relation to the road in a road video was observed by remote connection to the computer without connecting the camera. The code was then executed according to the images taken from the camera. In both structures, the vehicle has been shown to proceed in a desired manner. This study was also carried out as a graduation project and successfully completed.

## 5. THE PROPOSED METHOD

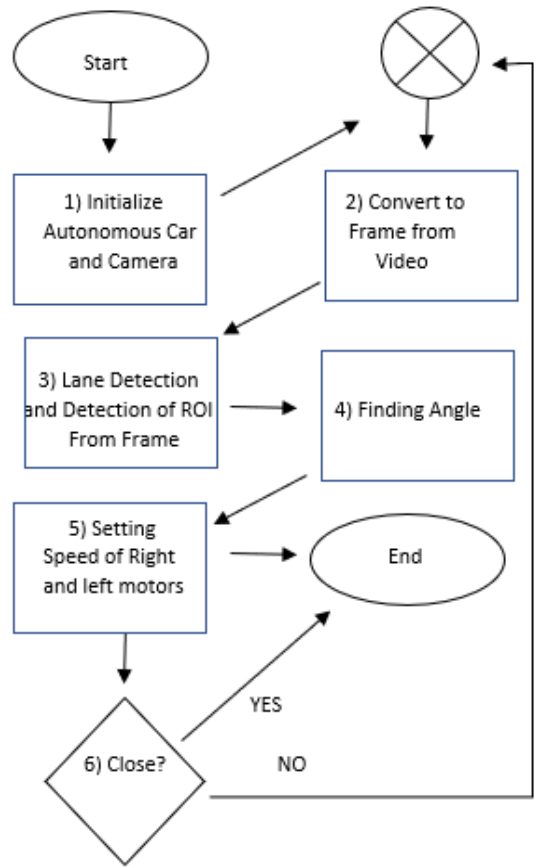

Algorithm. Pseudocode for the proposed method

```
1: // autonomous car and camera setting
2: while !CloseCar() do
3:         convert to frame from video
4:         detect to lane and ROI from frame
5:         find to angle of load
6:         set to speed of motor
7: end while
```

Figure 4. The proposed method: The Flowchart and The Pseudocode.

The figure above shows the algorithm-based operation designed in the period from the start of the vehicle to the closing of the vehicle.

When the vehicle is started, the camera also operates (Scheme1) and instant frames (frames) are taken from the camera (Scheme2). Lanes are detected by image processing methods and ROI is found (Scheme3). When the ROI is found, it is understood how the vehicle should be directed to protect the lane and there is an angle for this (Scheme 4). The angle found is adjusted by the speed of the right and left motors (Figure 5). As long as the vehicle is not switched off, the camera operates and returns to Scheme2, resulting in a flow loop. If the vehicle is switched off (Scheme6 takes yes), the flow switches to Scheme7 and the autonomous vehicle and camera are switched off.

## 6. CONCLUSION

In this paper, an autonomous vehicle prototype that detect lanes via image processing techniques, which are a major part of autonomous vehicle technology is presented. Some image processing algorithms such as canny edge detection, Sobel filter, etc. are used to provide autonomous movement capability. They were implemented and tested on our prototype vehicle. The prototype vehicle detected and followed the determined lanes successfully and it reached the destination. As a future work, we are planning to use generative neural networks, deep learning, and various machine learning algorithms to detect lane and traffic signs together.

## ACKNOWLEDGMENT

The project in this paper is presented as a bachelor thesis and it is supported by Istanbul University-Cerrahpasa Scientific Research Projects Commission as project number 32561 and "FBA-2019-33004". We would like to thank Istanbul University-Cerrahpasa Scientific Research Projects Unit for their support.

**Sertap Kamçı** completed her primary school education in Incirli Bahçe Primary School. She completed her secondary school education at the Siir Mektebi Primary School and his high school education at Gungoren Anatolian High School. She graduated Istanbul University-Cerrahpasa Faculty of Engineering Computer Engineering at 2019.

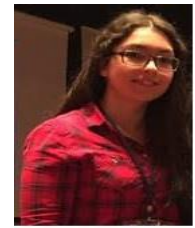

**Dogukan Aksu** received his B.S. and M.Sc. degrees in Computer Engineering at Istanbul University in 2015 and 2018 respectively. He is a PhD student in Computer Engineering. His research interests are information and network security, cryptography, artificial intelligence, machine learning and image processing.

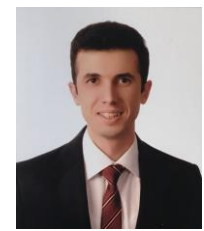

**Muhammed Ali Aydin** completed his B.S. in 1997-2001 in Computer Engineering at Istanbul University, M.Sc. in 2001-2005 in Computer Engineering at Istanbul Technical University and Ph.D. in 2005-2009 Computer Engineering at Istanbul University. His research interests are communication network protocols, network architecture, cryptography, information security and network security.

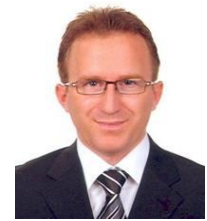